\documentclass{article}
\usepackage{spconf,amsmath,amssymb,graphicx}
\usepackage{booktabs}
\usepackage{bm}
\usepackage{url}
\usepackage[
  hidelinks, %
  ]{hyperref}

\newcommand{\ourmethod}{\mbox{PUAD}}

\title{\ourmethod: Frustratingly simple method for robust anomaly detection}

\twoauthors
 {Shota Sugawara\sthanks{This research was conducted as a summer internship project at LeapMind Inc.}}
	{The University of Tokyo, Tokyo, Japan\\
 poohshota@g.ecc.u-tokyo.ac.jp}
 {Ryuji Imamura}
	{LeapMind Inc., Tokyo, Japan\\
 rimamura@leapmind.io
 }
\begin{document}
\maketitle
\begin{abstract}
\def\thefootnote{\fnsymbol{footnote}}
Developing an accurate and fast anomaly detection model is an important task in real-time computer vision applications.
There has been much research to develop a single model that detects either structural or logical anomalies, which are inherently distinct.
The majority of the existing approaches implicitly assume that the anomaly can be represented by identifying the anomalous location.
However, we argue that logical anomalies, such as the wrong number of objects, can not be well-represented by the spatial feature maps and require an alternative approach.
In addition, we focused on the possibility of detecting logical anomalies by using an out-of-distribution detection approach on the feature space, which aggregates the spatial information of the feature map.
As a demonstration, we propose a method that incorporates a simple out-of-distribution detection method on the feature space against state-of-the-art reconstruction-based approaches. 
Despite the simplicity of our proposal, our method \ourmethod\footnote[2]{The source code is available at \url{https://github.com/LeapMind/PUAD}.}~(Picturable and Unpicturable Anomaly Detection) achieves state-of-the-art performance on the MVTec LOCO AD dataset.

\end{abstract}
\section{Introduction}
\label{sec:intro}
Image anomaly detection is a task to identify anomalous patterns within images, and it is often treated as a binary classification task to distinguish between normal and anomalous instances.
In recent years, there has been much research on anomaly detection methods using machine learning, with applications expected across a wide range of fields such as medical imaging~\cite{zhou2020encoding,schlegl2019f}, autonomous driving~\cite{ lis2019detecting}, and industrial inspection~\cite{cohen2020sub,8954181,bergmann2022beyond}.

The majority of the anomaly detection models rely on unsupervised learning due to the rarity and diversity of anomalous samples in practical applications.
The major anomaly detection models include feature-based~\cite{DBLP:journals/corr/abs-2005-14140,Roth_2022_CVPR,cohen2023set,DBLP:journals/corr/abs-2011-08785}, segmentation-based~\cite{liu2023component}, and reconstruction-based methods~\cite{Bergmann_2020,bergmann2022beyond, batzner2023efficientad}.
The feature-based method is a kind of out-of-distribution detection method, and it involves projecting images into a feature space using pre-trained feature extractors, with anomalies being assessed using either distance-based or density-based metrics.
The segmentation-based method initially divides images into various components using an unsupervised semantic segmentation model, subsequently classifying each element of the sample as normal or anomalous.
The reconstruction-based method determines the criterion for anomalies based on the distance of the sample's reconstruction error.
Based on these methods, various models have been developed. However, implementing these models in real-world situations remains difficult, entailing two critical factors.

\begin{figure}[t]
    \centering
    \centerline{\includegraphics[width=\linewidth]{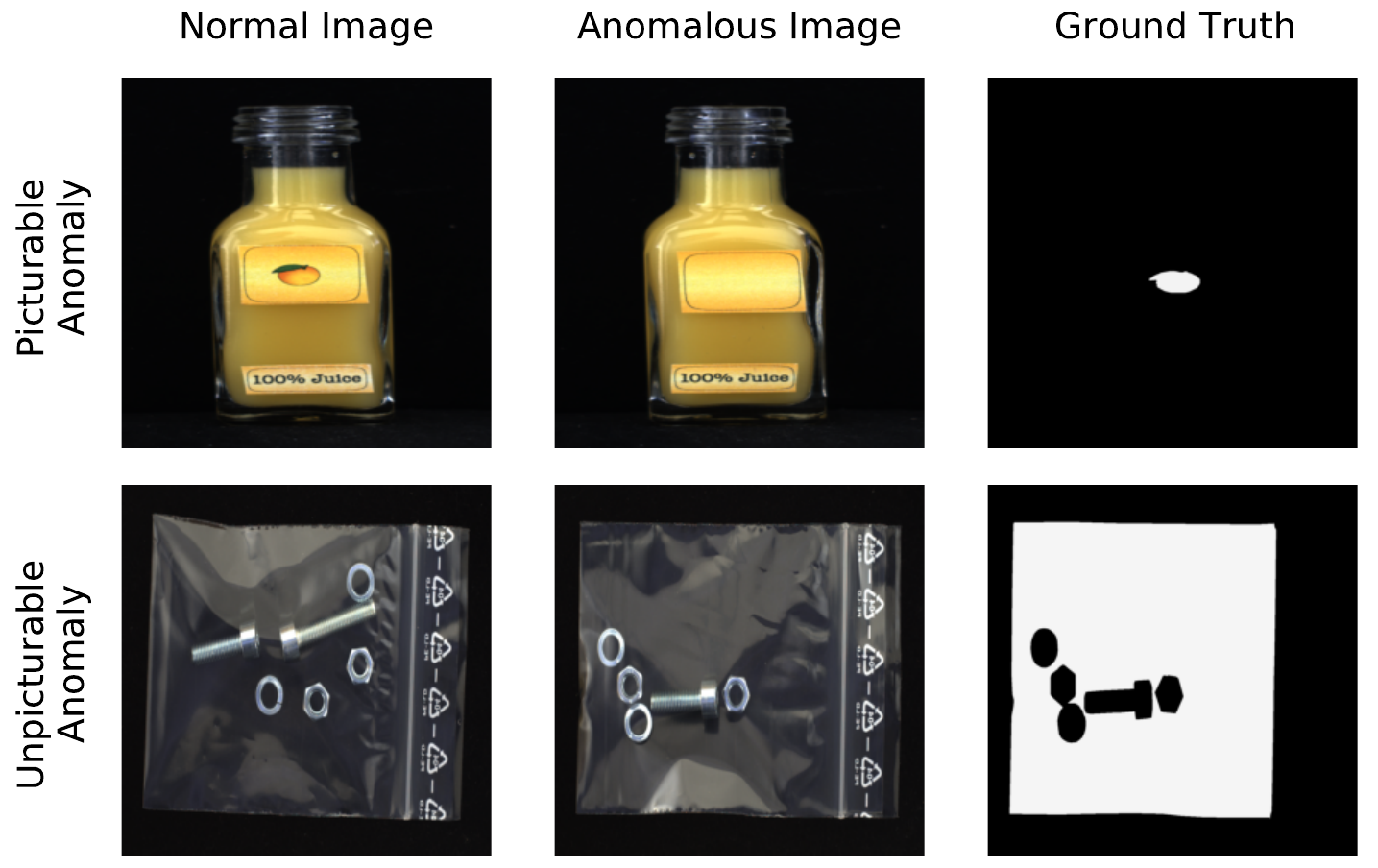}}
    \caption{
    These images from MVTec LOCO Dataset are examples of \emph{picturable} and \emph{unpicturable} anomalies.
    The white areas in the right column represent the Ground Truth, indicating the locations where anomalies are present.
    In the image of \emph{unpicturable} anomaly, the entire bag is marked as the ground truth, which means that the location of the anomaly is not clearly specified.
    }
    \label{fig:page1}
\end{figure}

Firstly, most existing models primarily focus on structural anomalies such as dents and scratches but face difficulties in detecting logical anomalies, such as missing components or incorrect quantities of components.
The recently introduced MVTec LOCO dataset~\cite{bergmann2022beyond} classifies anomalies into structural and logical categories, enabling the evaluation of anomaly detection methods' performance against logical anomalies.
The effectiveness of existing techniques on the dataset indicates considerable room for improvement.

Secondly, anomaly detection models often face a trade-off between inference time and anomaly detection accuracy.
Common approaches involve techniques such as model ensembling and utilizing large backbone networks.
Although these techniques improve the accuracy, the inference time usually becomes longer.
In real-world applications of anomaly detection, computation time limitations often need to be considered due to high throughput.
Therefore, considering both the inference speed and the model performance of anomaly detection methods is crucial to maintaining their applicability in practical settings.

Several methods address one of the two critical factors.
For example, the ensemble of heterogeneous models such as ComAD~\cite{liu2023component} and PatchCore~\cite{Roth_2022_CVPR} is a common technique to utilize their logical/structural detection capacity.
However, the naive model ensemble imposes an increase in the inference time.
Moreover, ComAD specializes in logical anomalies like quantity or number of objects, posing the issue of not being able to address anomalies that the method does not take into consideration, such as the wrong placements and separators coming off.
For another example, EfficientAD~\cite{batzner2023efficientad} states that it can detect both logical and structural anomalies with rapid inference speed, and it has a state-of-the-art score in the MVTec LOCO Dataset.
However, compared to the well-researched MVTec AD Dataset, the MVTec LOCO Dataset still has room for improvement in anomaly detection performance, especially with logical anomaly detection performance being lower than structural anomaly detection.

In this paper, prior to the model proposal, we qualitatively investigate the anomaly maps produced from EfficientAD, the current state-of-the-art model for the MVTec LOCO dataset. 
For some certain logically anomalous cases, we make consistent observations that EfficientAD completely fails to output the anomalous location on the anomaly map even though it is capable of detecting most of the anomaly types. 
Based on this observation, we consider that some logical anomaly types are not suitable to be represented via anomaly maps, and we call them \emph{unpicturable} anomalies.
To define it in more detail, we classify anomalies into two types: \emph{picturable} anomalies that can be represented by anomaly maps and \emph{unpicturable} anomalies that cannot be represented by anomaly maps.
In the case of the MVTec LOCO Dataset, all structural anomalies belong to \emph{picturable} anomalies since the locations of structural anomalies can be clearly illustrated on the anomaly maps.
On the other hand, logical anomalies include not only \emph{picturable} anomalies but also \emph{unpicturable} anomalies.

Figure~\ref{fig:page1} shows examples of anomalies belonging to the logical anomaly category in the MVTec LOCO dataset.
The top row in Figure~\ref{fig:page1} is an example of a \emph{picturable} anomaly, demonstrating that the location of the anomaly can be clearly depicted on the ground truth map.
The bottom row in Figure~\ref{fig:page1} is an example of \emph{unpicturable} anomaly.
The absence of a single screw is an anomaly, yet the entire bag is treated as the ground truth, making it impossible to identify the location of the anomaly uniquely.
It is impossible even for humans to decide where on the heatmap to illustrate the missing screw.
Existing reconstruction-based methods, which calculate an anomaly score from an anomaly map, may inherently struggle to detect \emph{unpicturable} anomalies.

Based on this insight, we propose a model that incorporates a simple feature-based detection method into EfficientAD, an existing reconstruction-based method.
Our approach, \ourmethod~(Picturable and Unpicturable Anomaly Detection), seeks to identify the \emph{picturable} anomalies with EfficientAD and the \emph{unpicturable} anomalies through a feature-based method.
Experiments using the proposed method demonstrated a performance 4.1 points higher than the baseline method, our implementation of EfficientAD.
The proposed method not only maintains stable high accuracy for various anomalies but also benefits from fast inference speed, as it does not use any network other than the one proposed in EfficientAD.
Furthermore, the essence of this study's idea is to apply different methodological approaches to \emph{picturable} anomalies and \emph{unpicturable} anomalies, which is not only applicable to reconstruction-based methods other than EfficientAD but also enables the replacement of the feature-based method with a more accurate one.

Our contributions are summarized as follows:
\begin{itemize}
    \item We analyzed the types of anomalies that existing state-of-the-art anomaly detection methods cannot detect and proposed to classify anomalies into \emph{picturable} anomalies, which are easy to detect by existing methods, and \emph{unpicturable} anomalies, which are difficult to detect.
    \item We proposed a new method that ensembles a feature-based method for \emph{unpicturable} anomalies and a \\reconstruction-based method for \emph{picturable} anomalies.
    \item We succeeded in improving the state-of-the-art for the MVTec LOCO dataset with rapid inference time.
\end{itemize}

\section{Related Work}
\label{sec:related_works}

\subsection{Anomaly Detection Tasks}
Among the various datasets, the MVTec AD dataset~\cite{8954181} is one of the most famous and often used as a benchmark for image anomaly detection.
The training set contains only normal images, and there are over 70 types of structural anomalies, such as scratches, dents, and contaminations, in the test set.

Recently, a new industrial anomaly detection dataset has been introduced called MVTec Logical Constraints (MVTec LOCO) dataset~\cite{bergmann2022beyond}.
This dataset comprises five categories.
Each category has a training set and validation set, which only contain normal images, and a test set, which contains both normal and anomalous images.
The novelty and noteworthy feature of this dataset is that it contains not only structural anomalies but also logical anomalies, such as a wrong ordering and a wrong combination of normal objects.
The MVTec LOCO dataset revealed that conventional models are good at detecting structural anomalies but weak at detecting logical anomalies.
We propose a robust anomaly detection model that can deal with both structural and logical anomalies.

\begin{figure}[t]
\begin{center}
\includegraphics[width=\linewidth]{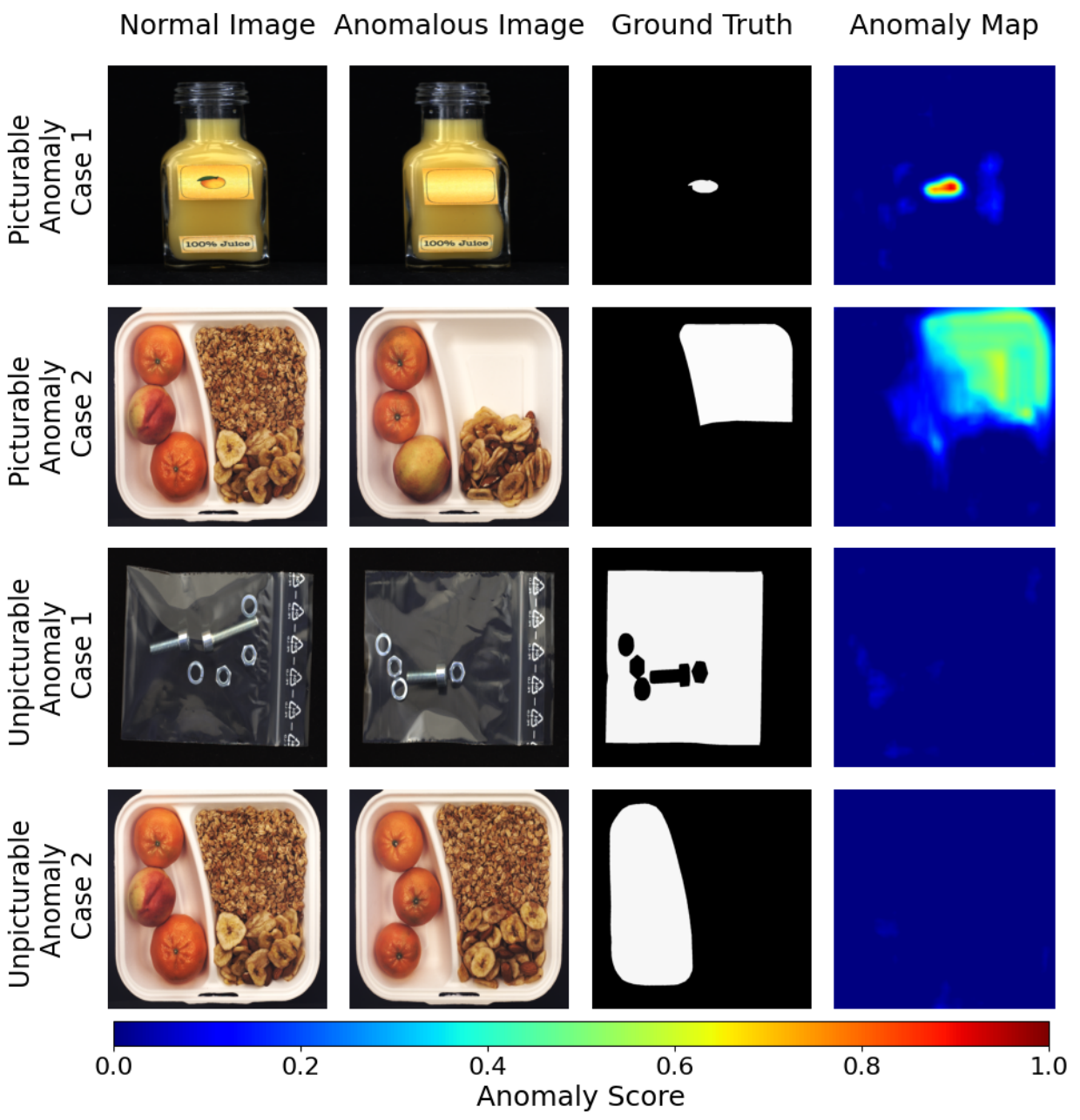}
    \caption{Normal images, anomalous images, ground truth maps, and anomaly maps of EfficientAD for \emph{picturable} and \emph{unpicturable} anomalies.
    In \emph{unpicturable} anomaly cases, the ground truth identifies all possible areas as candidates for the anomaly since it is impossible to identify a specific point of anomaly on the anomaly map.
    However, EfficientAD has not been able to produce the expected detection results.
    }
    \label{fig:unpicturable}
\end{center}
\end{figure}

\subsection{Anomaly Detection Methods}
Most of the existing methods can be classified into feature, segmentation, and reconstruction-based methods.

{\bf Feature-based methods} project the images into a feature space, and the degree of anomaly is quantified as the distance or density-based metrics.
In recent studies, pre-trained feature extractors have been widely used for projection.

GaussianAD~\cite{DBLP:journals/corr/abs-2005-14140} trains classification networks on ImageNet using only normal data and fits a multivariate Gaussian to the deep feature representations obtained from these networks, then calculates the Mahalanobis distance as the anomaly score.
PatchCore~\cite{Roth_2022_CVPR} stores normal samples as a feature bank and computes the distance between query features and the nearest neighbor key features as the anomaly criterion.
It achieves state-of-the-art results on the MVTec AD Dataset.
However, even though it uses a reduced database of clustered feature vectors to reduce inference time, searching for nearest neighbors using kNN during inference is very time-consuming.
In SINBAD~\cite{cohen2023set}, it introduces a technique for identifying anomalies that arise from unusual combinations of normal elements based on GaussianAD.
The model performs well on the MVTec LOCO dataset, suggesting that GaussianAD is also applicable to logical anomaly detection.

{\bf Segmentation-based methods} first segment images into multiple components based on an unsupervised semantic segmentation model, then classify each element of the sample as normal or anomalous or apply additional anomaly detection methods to each element.
ComAD~\cite{liu2023component}, specializing in logical anomaly detection using segmentation-based methods, shows high accuracy even for datasets containing both structural and logical anomalies by ensembling with a model that is good at detecting structural anomalies.
However, segmenting objects during inference tends to be time-consuming.

{\bf Reconstruction-based methods} are based on the assumption that models trained with normal samples are capable of reconstructing only normal patterns, leading to significant reconstruction errors in regions with abnormalities.
Reconstruction-based methods that use an autoencoder and GAN have been studied, and various models have been devised~\cite{Bergmann_2019,zavrtanik2021reconstruction,DBLP:journals/corr/abs-2108-07610,liu2022reconstruction,10106037,DBLP:journals/corr/SchleglSWSL17}.
Bergmann et al.~\cite{Bergmann_2020} introduces a Student-Teacher framework in which a collection of student networks is designed to match the local descriptors from pre-trained teacher networks using anomaly-free data.
This method identifies anomalies through increased regression error and predictive variance in the predictions made by the student network.
The networks used in this method have a restricted receptive field, limiting their ability to detect global inconsistencies that fall outside the receptive field's range.

GCAD~\cite{bergmann2022beyond} is designed to detect both structural and logical anomalies.
It has a local branch that uses the student-teacher method and a global branch that includes an autoencoder.
By combining the two branches, it is possible to detect both logical and structural anomalies.

\begin{figure*}[t]
\begin{center}
\includegraphics[width=0.97\linewidth]{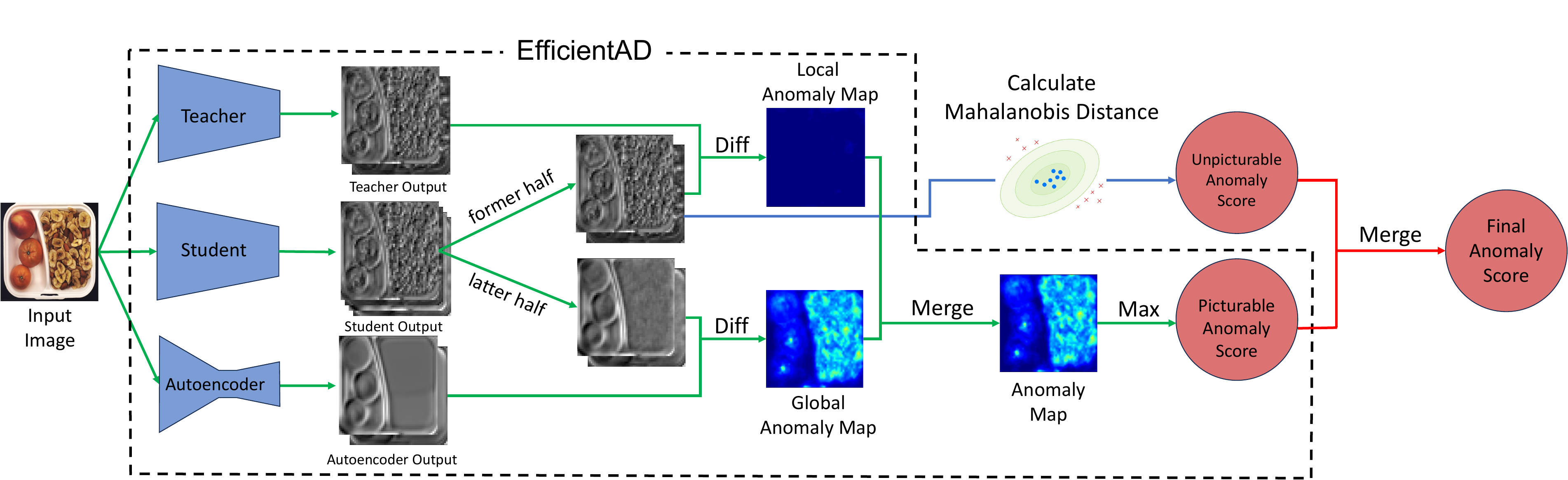}
    \caption{The scheme of the proposed method.
    We consider the score from EfficientAD as the \emph{picturable} anomaly score. 
    In addition, we consider the Mahalanobis distance calculated using the student's former half output or the teacher's output as the \emph{unpicturable} anomaly score.
    The final anomaly score is obtained by normalizing and summing up each score.
    }
    \label{fig:all-flow}
\end{center}
\end{figure*}

EfficientAD~\cite{batzner2023efficientad}, which is the state-of-the-art model on the MVTec LOCO dataset, inherits the idea of GCAD~\cite{bergmann2022beyond} that 
combining two separate convolutional neural network~(CNN) network modules for local anomalies and logical anomalies.
EfficientAD introduced a novel method that includes a patch description network~(PDN) that extracts local features, a model for structural anomaly detection using a special loss function, an autoencoder network for logical anomaly detection, and an excellent anomaly map normalization method.
Even though EfficientAD has high inference accuracy on average, the accuracy in some categories is significantly lower than that in other categories, leaving room for improvement.

\section{Proposed Method}
\label{sec:method}

\subsection{\emph{Picturable} and \emph{Unpicturable} Anomalies}
Most of the anomalies can be detected using the current state-of-the-art method, EfficientAD~\cite{batzner2023efficientad}.
Since it can detect not only structural anomalies but also some logical anomalies, the range of detectable anomalies has greatly expanded.
The images in the top two rows in Figure~\ref{fig:unpicturable} are examples of successful anomaly detection by EfficientAD.
The locations of anomalies are clearly displayed on the anomaly map, so we can easily identify them at a glance.

The cases in the bottom two rows of Figure~\ref{fig:unpicturable} show examples of applying EfficientAD to \emph{unpicturable} anomalous images.
Case 1 is considered an \emph{unpicturable} anomaly because it is not possible to uniquely determine where to draw the anomaly location for the missing screw within the bag.
In case 2, it is considered a normal image as long as there are two oranges and one apple in the breakfast box, regardless of their arrangement.
However, since the anomalous image contains three oranges and it cannot be uniquely determined which orange should be replaced with an apple, Case 2 is an \emph{unpicturable} anomaly.
In the ground truth, all possible areas are considered candidates for the anomaly location, yet in this instance, EfficientAD has completely failed to detect the anomalies.
Consistent with our hypothesis, it was confirmed that EfficientAD, a reconstruction-based method that calculates anomaly scores from anomaly maps, is capable of detecting \emph{picturable} anomalies but fails to detect \emph{unpicturable} anomalies.
Based on these results, our proposed method of detecting \emph{picturable} anomalies with a reconstruction-based method and \emph{unpicturable} anomalies with a feature-based method seems to be well-founded.

\subsection{\emph{Picturable} Anomaly Detection}
Except for detecting \emph{unpicturable} anomalies, EfficientAD has sufficiently high detection performance, as described in the previous section.
Therefore, we use this method as an anomaly detection module specialized for detecting \emph{picturable} anomalies.
The structure, training method, and anomaly calculation method of EfficientAD in this method follow the original paper. 
In other words, any other reconstruction-based anomaly detection method could be adopted as modules since no special modifications are required to apply this method.
An input image for EfficientAD is fed into three CNNs: teacher, student, and autoencoder.
Since the number of channels in the student's output is two times larger than the other CNN's output, we split it into a former half and a latter half to obtain a total of four outputs of the same shape.
We obtain the local anomaly map by averaging the teacher's output and the student's former half output along the channel direction and then computing the difference.
Similarly, we obtain the global anomaly map by applying the same process to the student's latter half output and the autoencoder's output.
We merge these two maps using the method proposed in EfficientAD to obtain the anomaly map shown in Figure~\ref{fig:unpicturable}. 
Furthermore, we obtain the \emph{picturable} anomaly score by taking the maximum value of the anomaly map.

\subsection{\emph{Unpicturable} Anomaly Detection}
We detect \emph{unpicturable} anomalies by using a feature-based method in addition to the existing reconstruction-based method.
Figure~\ref{fig:all-flow} illustrates the overall flow of \ourmethod.

\begin{table*}
\caption{The AUROCs when changing EfficientAD's size and network output for the Mahalanobis distance in \ourmethod.}
\label{table:efficientad-size-model-output}
\begin{center}
    \begin{tabular}{ccccc}
    \toprule
       EfficientAD size  & Output for the Mahalanobis & AUROC~(all)  & AUROC~(logical) & AUROC~(structural)\\
       \midrule
       S & Student's former half output & \bm{$93.12$} & \bm{$92.01$} & 94.12 \\
       M & Student's former half output & 92.23 & 90.30 & 95.35 \\
       S & Teacher's output & 92.24 & 91.09 & 94.65 \\
       M & Teacher's output & 91.78 & 89.44 & \bm{$95.49$} \\
       \bottomrule
    \end{tabular}
\end{center}
\end{table*}

\begin{figure}
\begin{minipage}[b]{0.48\linewidth}
    \centering
    \centerline{\includegraphics[width=4.0cm]{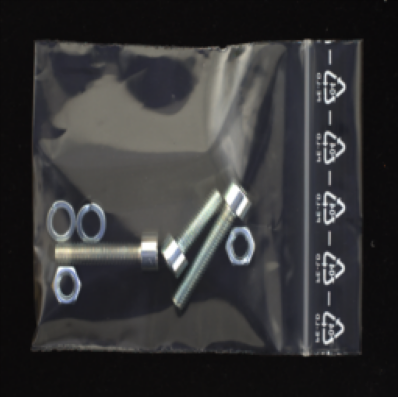}}
    \centerline{(a)}
\end{minipage}
\hfill
\begin{minipage}[b]{0.48\linewidth}
    \centering
    \centerline{\includegraphics[width=4.0cm]{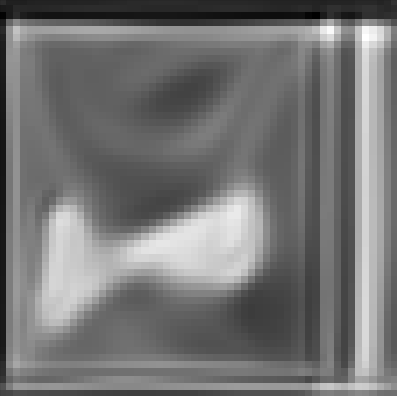}}
    \centerline{(b)}
\end{minipage}
\caption{(a) An original image in the screw bag category of the MVTec LOCO Dataset. (b) An autoencoder's output image in the screw bag category of the MVTec LOCO Dataset.\\
This shows that most features of logical information have been lost.}

\label{fig:autoencoder-output}
\end{figure}

GaussianAD~\cite{DBLP:journals/corr/abs-2005-14140} is one of the most famous and popular feature-based methods for anomaly detection.
GaussianAD generates feature vectors from the output of the middle layer of the deep feature extraction network and performs anomaly detection using Hotelling's T-square method.
We select GaussianAD as a feature-based method to integrate with EfficientAD because it is a very fast and simple feature-based method.

With reference to GaussianAD, we create feature vectors by applying global average pooling to the network output of EfficientAD.
Here, $\mathbf{x} \in \mathbb{R}^{H_{in}\times W_{in} \times C_{in}}$ is an input image, $\phi(\cdot):\mathbb{R}^{H_{in}\times W_{in} \times C_{in}} \rightarrow \mathbb{R}^{H_{out} \times W_{out} \times C_{out}}$ is either teacher network or student network, and $\mathbf{y}=\phi(\mathbf{x}) \in \mathbb{R}^{H_{out}\times W_{out} \times C_{out}}$ is the network output, where $H_{in}$,$W_{in}$ and $C_{in}$ are the number of elements in the vertical, horizontal, channel direction of the network's input and $H_{out}$,$W_{out}$ and $C_{out}$ are those of the network's output.
Letting the feature vector be $\mathbf{y}' \in \mathbb{R}^{C_{out}}$, the $c$-th element of $\mathbf{y}'$ can be calculated by
\begin{equation}
    \mathbf{y}'_c = \frac{1}{H_{out}W_{out}}\sum_{h=1}^{H_{out}}\sum_{w=1}^{W_{out}} \mathbf{y}_{c,h,w}.
\end{equation}
Then, we calculate mean $\boldsymbol{\mu} \in \mathbb{R}^{C_{out}}$ and convariance matrix $\mathbf{\Sigma} \in \mathbb{R}^{C_{out} \times C_{out}}$ of feature vectors with images in the train set.
The anomaly score can be calculated using the following formula in Hotelling's T-squared method:
\begin{equation}
    M(\mathbf{y}')=\sqrt{(\mathbf{y}'-\boldsymbol{\mu})^\top \boldsymbol{\Sigma} ^{-1}(\mathbf{y}'-\boldsymbol{\mu})}.
\end{equation}
$M(\cdot)$ is the Mahalanobis distance under a Gaussian distribution.

\begin{table}
\caption{Mean anomaly detection AUROC percentages of the MVTec LOCO Dataset.
The values are from the original paper except \ourmethod\ and EfficientAD~(Our Implementation)
}
\label{table:all-category}
\begin{center}
    \begin{tabular}{cc}
        \toprule
       Method  & AUROC \\
       \midrule
       GCAD~\cite{bergmann2022beyond}  & 83.3 \\
       ComAD~\cite{liu2023component}  & 81.2 \\
       ComAD + PatchCore~\cite{Roth_2022_CVPR} & 90.1 \\
       SINBAD~\cite{cohen2023set} & 86.8 \\
       EfficientAD~\cite{batzner2023efficientad}~(S, Our Implementation) & 89.0 \\
       EfficientAD~(M, Our Implementation) & 88.6 \\
       EfficientAD~(S) & 90.0 \\
       EfficientAD~(M) & 90.7 \\
       \ourmethod & \bm{$93.1$} \\
       \bottomrule
    \end{tabular}
\end{center}
\end{table}

There are three networks inside EfficinetAD: teacher, student, and autoencoder.
We need to consider which network and which layer outputs to use when calculating the Mahalanobis distance.
It may seem like a good idea to use the output or intermediate representation of the autoencoder because it was introduced to detect logical anomalies in EfficientAD.
However, there is a serious concern with this idea.
Figure~\ref{fig:autoencoder-output} presents an example of an output image of an autoencoder.
The abnormality in this image is the number of screws, and to detect the abnormality, it is necessary to recognize that the washers, nuts, and screws are separate parts, but this information is missing in the autoencoder's output.
Although the autoencoder can detect global anomalies, it is not suitable for detecting logical anomalies such as the number of parts, which EfficientAD is weak at.
For the same reason, the student's latter half output, which is trained to imitate the autoencoder's output, is not considered suitable for this purpose.
Therefore, we calculate the Mahalanobis distance using either the teacher's output or the student's former half output.

\subsection{Anomaly Score Normalization}

\begin{table*}
\caption{AUROC percentages by category and latency of EfficientAD and \ourmethod\ of the MVTec LOCO Dataset.}
\label{table:each-category}
\begin{center}
    \begin{tabular}{cccccccc}
    \toprule
       Method  & breakfast box & juice bottle & pushpins & screw bag & splicing connectors & mean &Latency~$\mathrm{[ms]}$ \\
       \midrule
       EfficientAD & 84.64 & 97.89 & 96.84 & 69.36 & 96.33 & 89.01 & \bm{$2.88$} \\
        \ourmethod & \bm{$87.07$} & \bm{$99.68$} & \bm{$98.02$} & \bm{$81.07$} & \bm{$96.76$} & \bm{$93.12$} & 3.00\\
        \bottomrule
    \end{tabular}
\end{center}
\end{table*}

Because the \emph{unpicturable} anomaly score is equivalent to the Mahalanobis distance and the \emph{picturable} anomaly score is normalized using the method proposed in EfficientAD, there isn't a common measure between them at the moment.
We normalize the \emph{picturable} and \emph{unpicturable} anomaly scores separately based on their mean and variance collected from the inference result of the validation subset.
The $\mathbf{x}_{z}$, which is the normalized $\mathbf{x}$, is calculated by  $\mathbf{x}_{z}=(\mathbf{x}-\boldsymbol{\mu})/{\sigma}$, 
where $\sigma$ is the standard deviation.

The final anomaly score is calculated as the sum of normalized \emph{picturable} anomaly score and normalized \emph{unpicturable} anomaly score.

\section{Experiments}
\label{sec:experiments}

\ourmethod\ and EfficientAD~(Our Implementation) were implemented using PyTorch.
All hyperparameters are set to the values in the original paper to implement EfficientAD.

The accuracy of anomaly detection was measured using the area under the Receiver Operating Characteristic curve~(AUROC).
We assessed the inference speed by measuring the duration from inputting the image into the network to obtaining the anomaly score as output.

Table~\ref{table:efficientad-size-model-output} reports the AUROC for each EfficientAD's size and network output used when calculating the Mahalanobis distance.
We can confirm that using the teacher network's output in EfficientAD is better at detecting structural anomalies but worse at detecting logical anomalies.
As Rippel et al.~\cite{DBLP:journals/corr/abs-2005-14140} mention, to create a model with high anomaly detection performance using GaussianAD, the feature extractor preferably has acquired the projection by training on various general images.
The characteristic of the student model, trained to reduce the norm of outputs to zero for inputs other than normal images, is presumed to be counterproductive when calculating the Mahalanobis distance for anomaly detection.
For logical anomalies, the student model shares weights with the model, which is trained to imitate the autoencoder's output, and that may play a part in improving accuracy since it can also address more global anomalies.

We can also confirm that increasing the size of the patch description network in EfficientAD leads to higher structural anomaly detection accuracy and lower logical anomaly detection accuracy.
We will not discuss it in depth because this result aligns with the tendencies of EfficientAD alone.
These results show that our proposal outperforms conventional methods in all combinations, although there are performance differences for logical and structural anomalies depending on which output is used.
Hereinafter, we use the score of the highest scoring combination, EfficentAD~(S) + student's former half output, as the score of the proposed method.

Table~\ref{table:all-category} presents the overall anomaly detection performance for each method.
\ourmethod\ uses our implementation EfficinetAD~(S) for \emph{picturable} anomaly detection and the student's former half output for \emph{unpicturable} anomaly detection.
\ourmethod\ achieves state-of-the-art performance in the MVTec LOCO Dataset, which means \ourmethod\ is good at detecting both structural and logical anomalies.

Table~\ref{table:each-category} provides the AUROC percentages by category of EfficientAD and \ourmethod.
We can confirm significant improvements in accuracy in the categories of breakfast box and screw bag.
These two categories, particularly the screw bag category, include more \emph{unpicturable} anomalies than other categories.
This result supports our hypothesis regarding \emph{unpicturable} anomalies.
Table~\ref{table:each-category} also provides the inference speed of EfficientAD and \ourmethod.
The results show that our algorithm is only about $0.12~\mathrm{ms}$, or roughly 5~\%, slower than EfficientAD since it uses a simple and fast feature-based method instead of using the additional network.
These results show that \ourmethod\ achieves significant performance gains while minimizing speed degradation.

\section{Conclusion}
\label{sec:conclusion}
In this paper, we presented a novel approach that combines robust anomaly detection capabilities with high computational efficiency.
We categorized anomalies into \emph{picturable} and \emph{unpicturable} anomalies and pointed out that the conventional reconstruction-based methods may not always be suitable for \emph{unpicturable} anomalies.
Our proposed method extends EfficientAD by introducing a feature-based method for detecting \emph{unpicturable} anomalies.
Compared to EfficientAD, we set a new state-of-the-art performance benchmark on the MVTec LOCO Dataset by increasing 4.1 points in the AUROC, while the inference speed becomes only $0.12\,\mathrm{ms}$ slower, which was less than a 5~\% change.
The essence of this proposal is that different approaches are necessary for \emph{picturable} and \emph{unpicturable} anomalies.
The results of this study suggest that rather than creating a single model capable of detecting all anomalies, models specialized for specific types of anomalies can improve the overall anomaly detection performance.
This indicates one valuable direction for future anomaly detection research.

\section*{Acknowledgments}
We would like to thank Yoshitaka Ushiku of NINE BULLS, Shuntaro Takahashi, Wasin Kalintha, and Joel Nicholls of LeapMind for their careful feedback on the manuscript.

\bibliographystyle{IEEEbib}
\bibliography{strings,refs}

\end{document}